\documentclass{article}

\usepackage{PRIMEarxiv}

\usepackage[utf8]{inputenc} 
\usepackage[T1]{fontenc}    
\usepackage{hyperref}       
\usepackage{url}            
\usepackage{booktabs}       
\usepackage{amsfonts}       
\usepackage{nicefrac}       
\usepackage{microtype}      
\usepackage{lipsum}
\usepackage{graphicx}
\graphicspath{{media/}}     

\usepackage{acro}
\DeclareAcronym{llm}{short=LLM, long=Large Language Model}
\DeclareAcronym{kg}{short=KG, long=Knowledge Graph}
\DeclareAcronym{re}{short=RE, long=Relation Extraction}
\DeclareAcronym{cot}{short=CoT, long=Chain-of-Thought}

\title{Construction of Hyper-Relational Knowledge Graphs Using Pre-Trained Large Language Models
}

\author{
  Preetha Datta\\
  Department of Computer Science, Aalto University\\
  Finland\\
  \texttt{preetha.datta@aalto.fi}
  \And
  Fedor Vitiugin\\
  Department of Computer Science, Aalto University\\
  Finland\\
  \texttt{fedor.vitiugin@aalto.fi}
  \AND
  Anastasiia Chizhikova\\
  Department of Computer Science, Aalto University\\
  Finland\\
  \texttt{anastasiia.chizhikova@aalto.fi} 
  \And
  Nitin Sawhney\\
  Department of Computer Science, Aalto University\\
  Finland\\
  \texttt{nitin.sawhney@aalto.fi}
}

\begin{document}

\maketitle

\begin{abstract}

Extracting hyper-relations is crucial for constructing comprehensive knowledge graphs, but there are limited supervised methods available for this task. To address this gap, we introduce a zero-shot prompt-based method using OpenAI's GPT-3.5 model for extracting hyper-relational knowledge from text. Comparing our model with a baseline, we achieved promising results, with a recall of 0.77. Although our precision is currently lower, a detailed analysis of the model outputs has uncovered potential pathways for future research in this area.

\end{abstract}

\section{Introduction}

\acp{kg} have captured widespread interest in both industry and academia, particularly in situations demanding the utilization of vast, constantly changing collections of data~\cite{hogan2021knowledge}. 
\acp{kg} use a graph-based data model to capture knowledge in application scenarios that involve integrating, managing and extracting value from diverse sources of data at large scale~\cite{nonaka1995knowledge}. 
Graphs offer a concise and intuitive representation for numerous domains, where connections and routes delineate diverse and sometimes intricate relationships among domain entities~\cite{ji2021survey}. 
Moreover, \acp{kg} enable maintainers to postpone the definition of a schema, allowing the data to evolve in a more flexible manner~\cite{angles2017foundations}.

\ac{re} task represents an opportunity for automated creation of extensive \acp{kg} by mining facts from natural language text. 
In general, these methods concentrate on binary relations~\cite{han2020more}, linking two entities to form a relation triple: the starting entity, the relationship, and the ending entity.
However, real \acp{re} often feature complex hyper-relational facts~\cite{guan2019link}.

\begin{figure}[!h]
    \centering
    \includegraphics[width=0.37\textwidth]{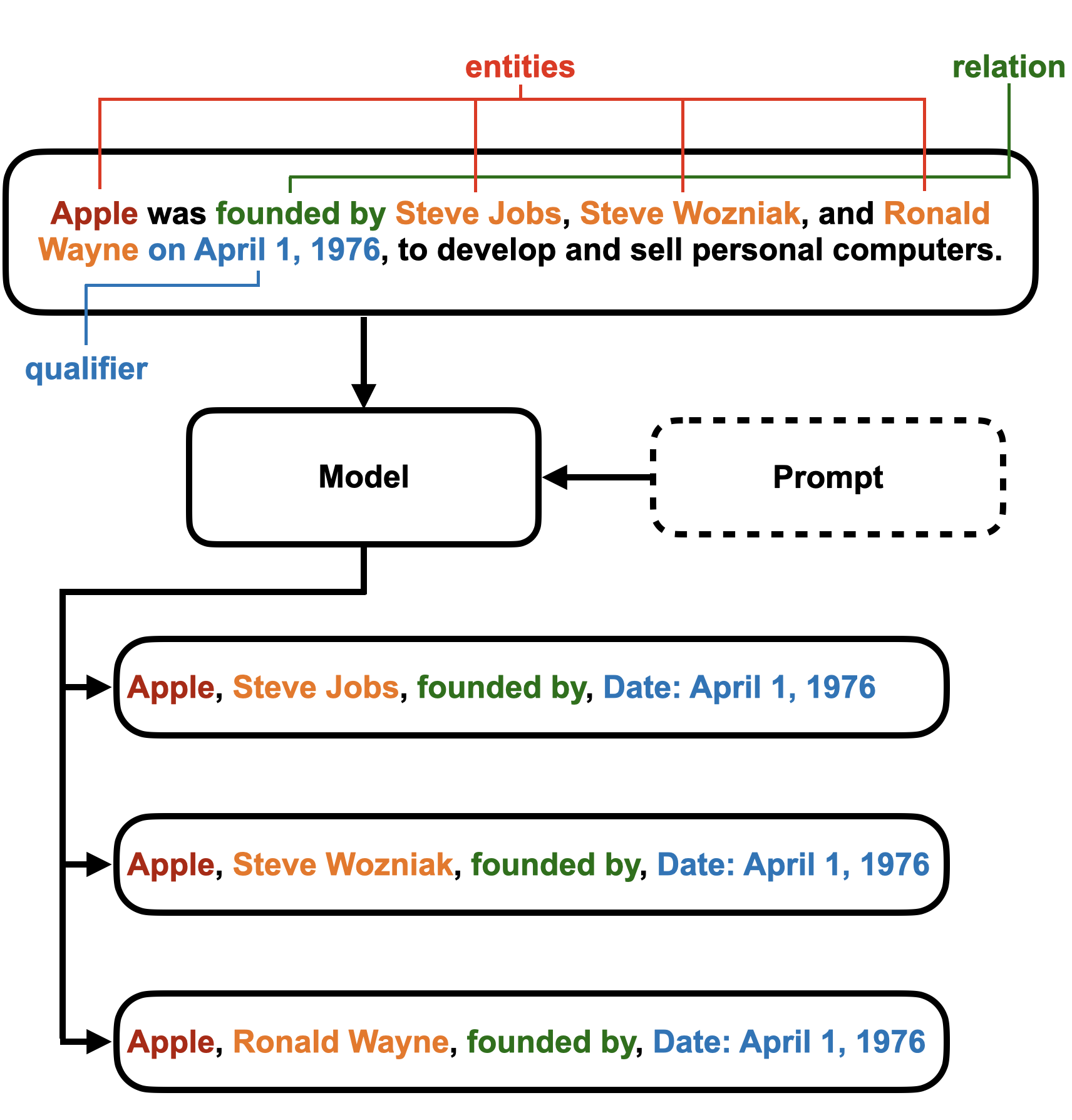}
    \caption{Example of extraction entities, relations and qualifiers from the text using our zero-shot prompting technique.} 
    \label{re_example}
\end{figure}

These facts include additional qualifier attributes -- like time, quantity, or location -- for each relationship triple, as illustrated in the provided Figure~\ref{re_example}. 
Therefore, simply extracting these relation triples might oversimplify the intricate structure of \acp{kg}, failing to capture its richness and complexity.

Hyper-relational facts differ from standard relation triples because their qualifiers apply to the entire triple rather than a specific entity within it.
Despite this complexity, hyper-relational facts offer practical advantages, enhancing fact verification~\cite{thorne2018fever} and refining how \acp{kg} are represented and learned~\cite{galkin-etal-2020-message}.
Consequently, extracting relation triples with their associated qualifiers becomes essential for constructing these hyper-relational facts. 
Unfortunately, there's a scarcity of models dedicated to hyper-relational extraction. 
For instance, the CubeRE model~\cite{chia-etal-2022-dataset}, a supervised learning approach, which requires a training dataset and presents uncertainties regarding its adaptability to different domains.

The key motivation behind our work is the promise of significant saving of resources requested for building training data. We found that leveraging \acp{llm} for information extraction could have noteworthy implications, especially in scenarios where adapting to specific domains or tasks is critical. The use of pre-trained \acp{llm} has the potential to expedite the deployment of information extraction techniques across diverse domains, offering an efficient and resource-friendly alternative to traditional supervised approaches.

One of the key contributions of our work is a prompt-based method for hyper-relational extraction from unstructured text data, which is used for construction of rich and complex \acp{kg}.

\section{Related Work}

The term ``Knowledge Graph'' was originally introduced by Google in a seminal blog post in 2012, signifying a ``critical first step towards the next generation of search''~\cite{Singhal_2012}. While graph-based textual documents have existed for decades, it was this blog post that largely popularized and highlighted the flexibility of the \ac{kg} architecture.

\subsection{Relation Extraction}

While \acp{kg} could be constructed based on semi-structured websites~\cite{sarkhel2023self}, tables~\cite{dong2020multi} or link prediction~\cite{rossi2021knowledge}, extracting facts from unstructured data is still the most challenging task due to the vast amount of text on the web~\cite{ye2022generative}.

\ac{re} is one of the main parts of \acp{kg} construction and a well-established task which could be done with supervised~\cite{tran-etal-2022-improving} or semi-supervised~\cite{han-etal-2019-opennre} learning methods. 
While the Zero-Shot Relation Extraction approach \cite{tran-etal-2022-improving} could adapt and extrapolates knowledge across diverse domains and applications, it faces limitations when labeled training data is scarce or expensive. In such cases, unsupervised techniques offer compelling alternatives by combining Pre-Trained Language Models such as RoBERTa, with Open Information Extraction~\cite{temperoni2022enriching}. 
Prompt-Based Zero-Shot Relation Extraction Method is a prominent technique to maximize the efficiency of data exploitation by prompt-tuning technique in pre-trained language models~\cite{jun-etal-2022-exploration}.

The use of carefully crafted instructions, known as prompts, allows to better uncover connections and details in unstructured textual data~\cite{hu2023zero}. 
A recent approach called ChatIE transforms the zero-shot information extraction method into a multi-turn question-answering framework by using prompt-like conversation with the \ac{llm} to construct a robust information extraction model~\cite{wei2023zero}

\subsection{Prompt Engineering in \acp{llm}}
Before discussing prompt engineering in detail, let's examine \acp{llm}, and more specifically, the GPT series of models released by OpenAI. The specific technical details, encompassing aspects such as architecture, hardware, dataset construction, and training methodology, remain undisclosed for the OpenAI's GPT models~\cite{10113601}. Overall, GPT models are autoregressive language models -- meaning it utilizes a decoder-only approach to language modeling, and employs a one-way language encoding-decoding method and token-by-token prediction of words.

Prompt engineering serves as an effective paradigm that closes the divides between pre-training tasks and their corresponding downstream applications in case of \acp{llm}~\cite{kan2022unified}. There have been several prompting techniques specifically explored for information extraction. One such technique is \ac{cot} prompting. \ac{cot} prompting~\cite{wei2022chain} is a technique that involves presenting a coherent series of intermediate reasoning steps as exemplars in the prompting process. Research indicates that employing \ac{cot} prompting enhances performance across various tasks, including arithmetic, commonsense, and symbolic reasoning. 

However, when working with \ac{llm}'s, one must always be cognizant of erroneous predictions or what is popularly referred to as "hallucinations". Within the NLP community, the term "hallucination" while problematic in many ways (since it gives agency to the AI system rather than its creators), has been extensively embraced; it commonly denotes the generation of information that deviates from logical or faithful representation of the given source content. This consideration is pivotal when navigating the intricacies of \acp{llm}, as the potential for generating nonsensical or inaccurate information necessitates a nuanced approach to model evaluation and refinement. There could be many reasons attributed to so-called hallucinations, whether it is incompleteness in the prompting technique, bias in the training data corpus, or under informativeness of the \ac{llm} if the question is particularly niche~\cite{zhang2023siren}. 

\subsection{Hyper-Relational Knowledge Graphs}

\begin{figure}[htb]
    \centering
    \includegraphics[width=0.37\textwidth]{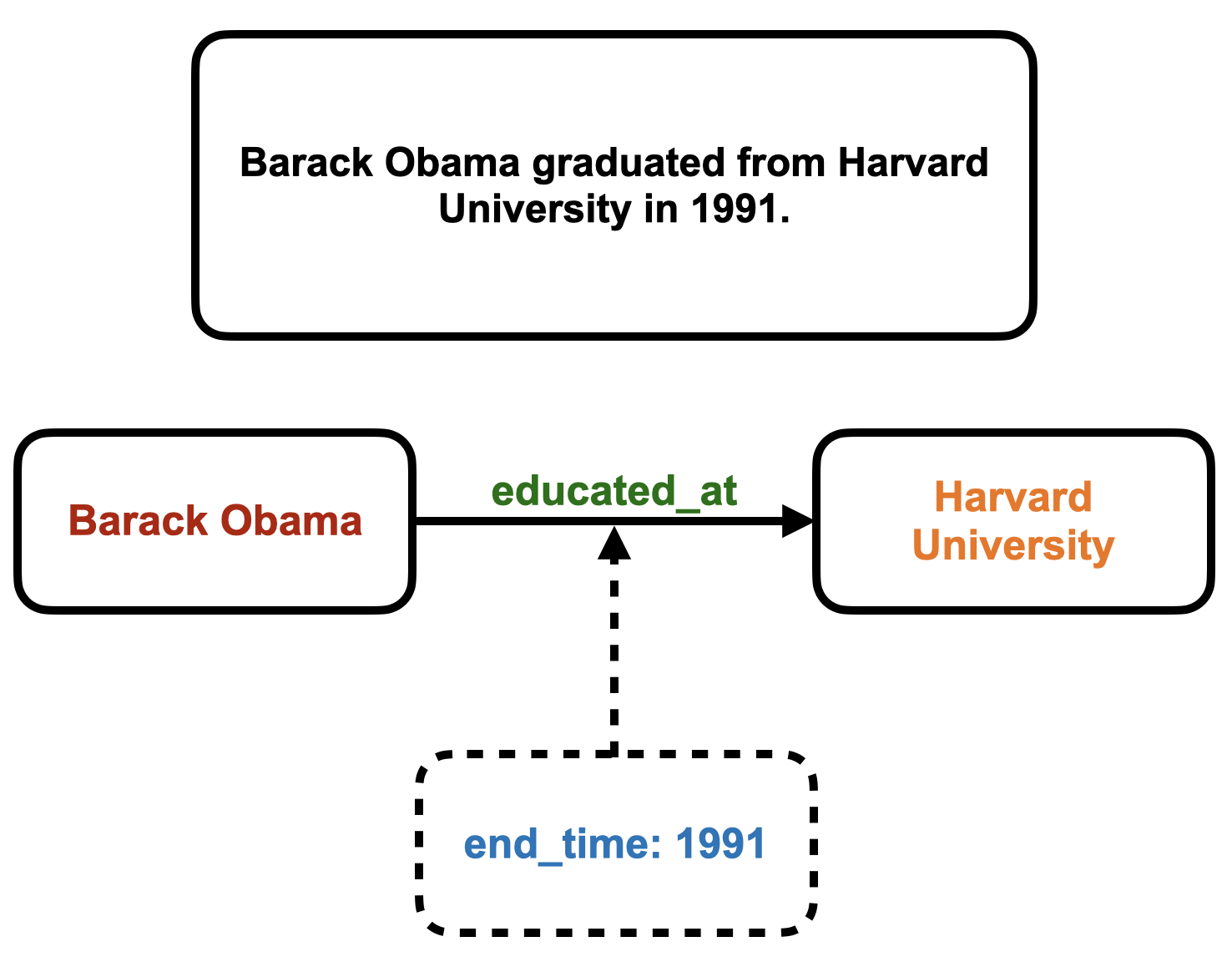}
    \caption{Example of Hyper-Relational Knowledge Graph.} 
    \label{hr-kg}
\end{figure}

While \acp{kg} are a great way to encapsulate simple, factual information, real data tends to differ with specific nuances, conditionals and additional information for extracted fact. 
A hyper-relational \acp{kg}, such as Wikidata, is a \acp{kg} that enable the inclusion of supplementary \textbf{key:value} pairs alongside the primary triple, offering a means to clarify or limit the applicability of a given piece of information~\cite{galkin-etal-2020-message}. Hyper-relational \acp{kg} can be understood as follows -- consider the sentence, \textit{``Barack Obama graduated from Harvard University in 1991.''}. In the context of conventional triple-based \acp{kg} it becomes evident that the richness of the information in the sentence is not fully encapsulated by mere triples. However, Figure~\ref{hr-kg}, which includes qualifiers as part of its hyper-relational \acp{kg} structure, is easily capable in showcasing the nuances embedded in the given sentence.

Our paper is providing the novel zero-shot prompt-based  approach for hyper-relational extraction from the HyperRED dataset~\cite{chia-etal-2022-dataset}.

\section{Methodology} 

This section provides an in-depth description of our prompting techniques in constructing \acp{kg} using \acp{llm}, along with details of method's implementation.

\subsection{Prompting Technique}

Prompt engineering is the process of designing and creating prompts that generate desired responses from language models.
These prompts can be classified into various types based on their structure, function, and complexity. Each prompt comprises a natural language query crafted to elicit a particular response from the pre-trained language model.

In our prompts to \ac{llm}, we incorporate an ontology presented in the HyperRED dataset, coupled with a \ac{cot} example to illustrate the desired outcome for the relation and entity extraction task. Additionally, we define an expected output format that we anticipate the model to follow. More specifically, our prompts instruct the model to extract relations, entities, and qualifiers from the at the time. We detail a description of \textit{entities}, describing it as a person, country, type of document etc. We also give a lengthy description of the 62 \textit{relations} according to HyperRED ontology. For example, we define the relation \textit{``educated at''} as \textit{``educational institution attended by subject''}. Finally, we describe what a \textit{qualifier} is, and supply to the model individual definitions of the 44 qualifiers available in the dataset's ontology. While the qualifiers themselves are expected in a \textbf{key:value} format, we supply the model with a detailed description of the set of qualifier descriptions in order for there to be better contextual understanding. For example, \textit{``end time''} is defined as \textit{``time an event ends, an item stops existing, or a statement becomes invalid''}. The \textbf{key:value} format qualifier pair can be seen in Figure \ref{hr-kg}. For example, a correct qualifier output would be as follows -- \textit{``end time: 1991''}. The motivation behind this structure was to prevent as much erroneous prediction of information by the \ac{llm} as possible. The final prompt used for our model is available in a public repository~\footnote{https://github.com/preetha-sellforte/CoT-GRex}.

Finally, to apply \ac{cot} prompting, we give the model one example of a context and clear reasoning as to why we choose to extract the specific entities, relations and qualifiers from that sentence. We also provide an output format that we expect the model to adhere to. 

\subsection{OpenAI's GPT-3.5}

We use the pre-trained OpenAI's GPT-3.5 (GPT-3.5) model and access it via Azure API calls. OpenAI's GPT-3 (Generative Pre-trained Transformer) model was first introduced by OpenAI~\cite{brown2020language} company in 2020. OpenAI's GPT series of models are transformer-based architectures~\cite{vaswani2017attention}, as the name might suggest. 

OpenAI's GPT-3 model underwent fine-tuning through reinforcement learning from human feedback~\cite{christiano2017deep}. This method leverages rewards to enhance the quality of generation and ensure alignment with human preferences. Several studies have been conducted researching the mathematical, reasoning, text generative properties of GPT-3.5~\cite{LOPEZESPEJEL2023100032}, which is why we have decided to use the model due to its widely regarded state-of-the-art status~\cite{ye2023comprehensive}.

\section{Experiment Setup}  

This section contains a detailed description of the dataset, baseline model and evaluation metrics used for our experiment.

\subsection{Dataset}

We utilized a hyper-relational dataset HyperRED spanning diverse domains such as business, sports, and politics. 
The dataset creation involved primarily collecting data from Wikipedia introduction sections, followed by entity linking and fusion using DBpedia Spotlight. As Wikipedia articles often tend to use pronouns when referring to a subject or entity, the authors also used the Stanford CoreNLP tool~\cite{manning-etal-2014-stanford} to resolve coreferences. 
The dataset's annotation by two professional annotators resulted in 39,840 training samples, 1,000 development samples, and 4,000 test samples. HyperRED comprises a set of 62 unique relations and 44 unique qualifiers. 
The dataset's authors declared it as the first-ever dataset with hyper-relational facts.

\subsection{Baseline Model}

Authors of the HyperRED dataset also present CubeRE, a supervised learning method aimed at extracting hyper-relational information.
CubeRE employs a two-step process for hyper-relational extraction. 
Firstly, input sentences are encoded using the BERT~\cite{devlin2018bert} language model, generating contextualized representations for each word. The model then captures interactions between potential head and tail entities by concatenating word representations and using a dimension-reducing feed-forward network. 
This results in a \textit{cubic} (3-dimensional) matrix of probabilities for entity and relation labels. 
Secondly, to predict qualifiers for hyper-relational facts, interactions are considered between possible relation triples and value entities. To manage computational costs, CubeRE introduces a cube-pruning method, focusing on interactions with the \textit{top-k} words with the highest entity scores. Entity scores are obtained from the diagonal of the 3D matrix, and the \textit{top-k} indices are ranked for each dimension of the cube representation. 

\subsection{Evaluation Metrics}

In addition to classic \textit{precision}, \textit{recall} and \textit{f1} evaluation metrics used for relation extraction task, we additionally utilized  BERTScore metric.
BERTScore~\cite{zhang2019bertscore} serves as an automated evaluation metric designed for the evaluation of text generation quality. We use BERTScore in our evaluations since it employs contextual learnings that can capture the semantic similarities of the information extracted by the \ac{llm}. This is useful in our case since we expect the model to capture more relations and qualifiers.

\section{Results}

Table \ref{exact-res} presents results for exact-match scores between the CubeRE method and the proposed approach. 
Initially, it seemed our model performed very poorly comparing to the baseline according to the exact match metric. 
However, during the prompt engineering step, we observed very promising results. To understand the reason of such unsatisfactory performance, we carried out an analysis of the generated relations.

\begin{table}[!ht]
\caption{Table encompassing the precision, recall and F1 scores on exact-match evaluations.}
\centering
\begin{tabular}{|l|l|l|l|}
\hline
\textbf{Model} & \textbf{Precision} & \textbf{Recall} & \textbf{F1} \\ \hline
GPT-3.5 & 0.01 & 0.02 & 0.01 \\ \hline
CubeRE (Baseline) &\textbf{0.62} & \textbf{0.66} & \textbf{0.64 }\\ \hline
\end{tabular}%

\label{exact-res}
\end{table}

\begin{figure}[!h]
    \centering
    \includegraphics[width=0.37\textwidth]{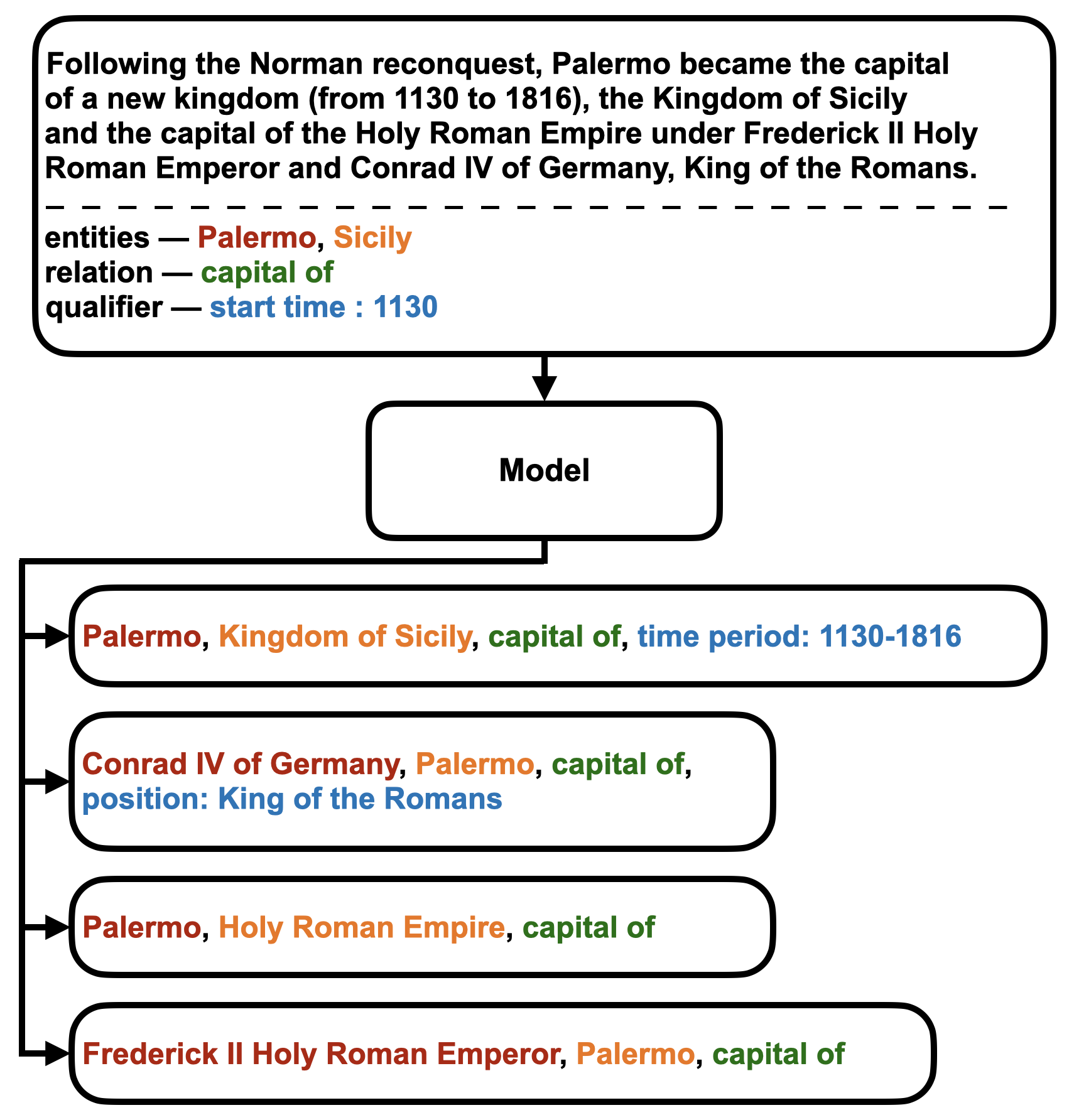}
    \caption{Comparing gold standard annotation, with our results. The model extracts information from the given sentence in high granularity, but often not in an ``exact'' format.} 
    \label{triplet-comp}
\end{figure}

Further exploration presented in Figure~\ref{triplet-comp} illustrates how the proposed method extracts entities and relations from source texts. The first hyper-relation is very close to the gold standard. The second hyper-relation is erroneous, by providing a mix of person (entity -- \textit{Conrad IV of Germany}, qualifier -- \textit{position: King of the Romans}) and place (entity -- \textit{Palermo}, relation -- \textit{capital of}). The third relation is a correct triple describing the capital relation between city and country. The last relation is an incorrect triple. In conclusion, we can see the proposed method extracts a correct hyper-relation, but it also extracts additional relations which could be incorrect. Therefore, we decided to explore non-exact match evaluation.

Results of the semantic similarities' evaluation made with BERTScore are presented in the Table~\ref{bertscore-res}. 
The promising recall score obtained during the experiment indicates the ability of the proposed approach to capture high proportion of hyper-relations. 
The results could be substantially enhanced with more complex prompting techniques, as well as fine-tuned model settings specific to the task at hand.
Additionally, some filtering  and ranking mechanisms applied to the model's output could also improve the precision. 

We also measured reproducibility of our approach with the use of Levenshtein Distance in several experimental outputs of the same data samples. The average reproducibility score during experiments with GPT-3.5 equals 36\%, while the reproducibility score during experiments with OpenAI's GPT-4 equals 98\%. Such a crucial increase in reproducibility became possible because of the support of predefined low randomization of OpenAI's GPT-4 model outputs (temperature = 0).

\begin{table}[!h]
\caption{Table encompassing the precision, recall and F1 scores on BERTScore evaluations.}
\centering
\begin{tabular}{|l|l|l|l|}
\hline
\textbf{Model} & \textbf{Precision} & \textbf{Recall} & \textbf{F1} \\ \hline
GPT-3.5 & 0.46 & 0.77 & 0.58 \\ \hline
CubeRE (Baseline) &\textbf{0.93} & \textbf{0.88} & \textbf{0.91 }\\ \hline
\end{tabular}
\label{bertscore-res}
\end{table}

\section{Conclusions \& Future Work}

In this paper, we present a zero-shot prompting technique for hyper-relational \acp{kg} construction using \acp{llm}. We have utilized the HyperRED dataset and compared our results against the cubeRE method.

Despite not surpassing the baseline, further exploration of sophisticated prompting techniques should yield better outputs from the model. Additionally, given the promising recall, attention should be directed toward enhancing precision for continued improvement. We should also look at fine-tuned settings in GPT models, as well as look for alternatives in open-source domains, such as Meta's Llama 2~\cite{touvron2023llama}. These measures could significantly improve our results. 

In our future work, we will explore different techniques of prompting, as recently outlined~\cite{sivarajkumar2023empirical}. We also plan to create an approach to filter out irrelevant information in the given context, to be able to improve the performance. 
Finally, we intend to test our approach on a close-domain dataset that contains detailed instructions and valuable information about migrants' employment in Finland. This experiment aims to demonstrate the flexibility and effectiveness of our method across different domains. 

\section*{Acknowledgments}
This work is supported by the Trust-M research project, a partnership between Aalto University, University of Helsinki, Tampere University, and the City of Espoo, funded in-part by a grant from the Strategic Research Council (SRC) in Finland.

\bibliographystyle{plain}  
\bibliography{references}

\end{document}